\documentclass[cameraready]{Interspeech}
\title{Tanawpamana: Benchmarking Vision-Language Models on Philippine Cultural Heritage Sites
}

\author[affiliation={1,2}, correspondingauthor]{Ivan Yuri}{De Leon}
\author[affiliation={3}, ]{Hiroki}{Ouchi}
\author[affiliation={3}, correspondingauthor]{Sakriani}{Sakti}

\address{
    $^1$ Ateneo de Manila University, Philippines \\
    $^2$ Globe Telecom, Philippines \\
    $^3$ Nara Institute of Science
and Technology, Japan
    }

\email{ivan.deleon@student.ateneo.edu, \{hiroki.ouchi,ssakti\}@is.naist.jp }


\usepackage{comment}

\usepackage{xcolor}
\usepackage{colortbl}
\usepackage{booktabs}

\definecolor{softblue}{HTML}{9DBFDA}
\definecolor{offwhite}{HTML}{F2F0EC}
\definecolor{highlightblue}{HTML}{2E5C8A}

\newcommand{\lo}[1]{\cellcolor{softblue}#1}

\begin{document}

\maketitle

\begin{abstract}
Vision-language models are increasingly used to identify and describe places, yet their recognition and knowledge of specific cultures remain poorly understood, particularly in underrepresented regions such as the Philippines. This study introduces the Tanawpamana Dataset, a repository of Philippine heritage site images with metadata based on the Talapamana registry of the National Commission for Culture and the Arts of the Philippines. We also introduce the Tanawpamana Benchmark, comprising 4,046 tasks designed to evaluate models' recognition and knowledge of heritage sites. Evaluating seven vision-language models, we find that they struggle more with fine-grained visual discrimination among similar heritage sites than with coarse recognition, and that their geographic knowledge is weaker when inferred from image inputs than from text. All resources are available on GitHub.\footnote{\url{https://github.com/thelionlies/tanawpamana}}
\end{abstract}

\section{Introduction}
\label{sec:intro}

\begin{figure}[!t]
    \centering
    \includegraphics[width=\linewidth]{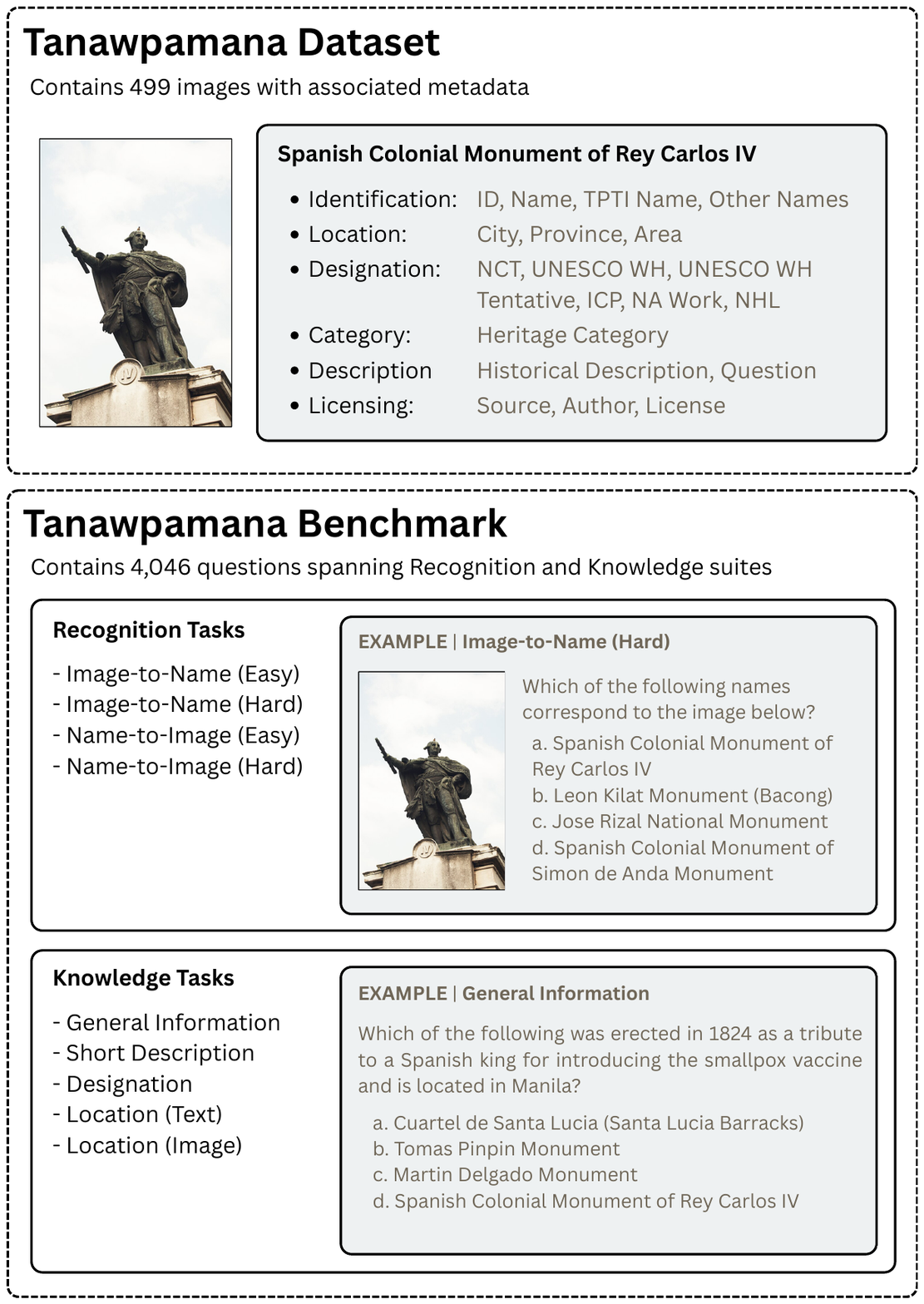}
    \caption{Overview of the Tanawpamana Dataset and Tanawpamana Benchmark. Top: sample dataset entry showing a heritage image and metadata fields. Bottom: Recognition and Knowledge task suites with simplified example questions.}
    \label{fig:tanawpamana_tasks}
\end{figure}

The Philippines is home to over 180 ethnolinguistic groups, with a complex history layered by pre-colonial practices and centuries of Spanish, American, and Japanese encounters. This diversity is reflected not only in the country's living traditions but also in its built environment: churches, forts, bridges, and monuments that each carry a unique history.

As people increasingly rely on artificial intelligence systems to access information about places and cultural heritage, the ability of these systems to recognize and represent diverse cultures becomes increasingly important. Vision-language models (VLMs), which jointly process images and text, are central to this question because they shape how images of places and cultural heritage are interpreted. Assessing whether these models understand a particular culture requires high-quality datasets and benchmarks that explicitly evaluate culturally grounded knowledge.

While a growing body of Southeast Asian datasets and benchmarks has begun to address regional underrepresentation, the Philippines remains largely absent from vision-language model evaluation, particularly for cultural heritage understanding. SEACrowd \cite{lovenia2024seacrowd} provides a broad resource landscape covering nearly a thousand Southeast Asian languages, but only a small fraction of its resources are image-based, limiting its applicability for evaluation that requires visual cultural representations. SEA-VL \cite{cahyawijaya2025crowdsource}, the largest such effort at approximately 1.28 million images, relies primarily on crowdsourced and automated data collection, with only 543 of its 8,018 human-verified images related to the Philippines. RICE-VL \cite{pranav2025rice} evaluates cultural understanding across eleven ASEAN countries, but reports only a single aggregate score per country across fourteen combined categories, preventing the recovery of Philippine-specific results. The Seeing Culture Benchmark \cite{satar2025seeing} expands cultural evaluation across seven countries through a shared collection of 1,065 images, but does not specifically target cultural landmarks or tangible heritage sites. SEA-VQA \cite{urailertprasert2024sea} provides the most explicit accounting of Philippine cultural content among these efforts, drawing from UNESCO's Cultural Heritage collection, but includes only 69 images spanning six Philippine cultural practices.

Across all five, the same pattern recurs: Philippine content is either aggregated beyond the point of isolating country-specific results, represented only marginally within broader multilingual resources, or limited by small-scale or weakly attributable image collections. None is designed to evaluate vision-language models on Philippine cultural contexts or to assess recognition and knowledge of tangible heritage sites. With these, it remains unclear whether these models can reliably recognize and retrieve information about Philippine heritage from visual and textual information.

To address this gap, we make the following contributions:

\begin{enumerate}
    \item We introduce the \textbf{Tanawpamana Dataset}, a collection of 499 publicly available images of Philippine cultural heritage sites sourced primarily from Wikimedia Commons and Philippine government repositories, with metadata enriched from the National Commission for Culture and the Arts' (NCCA) \textit{Talapamana} registry, including site names, locations, heritage categories, and official designations.
    \item We introduce the \textbf{Tanawpamana Benchmark}, a 4,046-item evaluation suite built on this dataset, comprising a recognition benchmark that isolates VLMs' ability to visually identify specific heritage site entities, and a knowledge benchmark that isolates the models' entity-level factual, geographic, and designation knowledge, including a paired image-versus-text test of geographic knowledge.
\end{enumerate}

\section{Related Work}

\subsection{Cultural and Vision-Language Benchmarks}

Vision-language model evaluation grew out of visual question answering (VQA), which paired images with natural-language questions and scored model responses against ground-truth answers \cite{antol2015vqa}. Later benchmarks expanded VQA toward more complex multimodal reasoning, as in MMMU's 11,500 college-level questions \cite{yue2024mmmu}, and other work has sought to improve the reasoning of vision-language models themselves, for example by fine-tuning on a small set of curated reasoning chains for knowledge-based VQA \cite{maulana2025limvo}.

Beyond general reasoning ability, a separate line of work asks what models actually recognize and know. OVEN \cite{hu2023open} distinguishes category-level recognition (e.g., a castle) from entity-level recognition (e.g., Windsor Castle), noting that many benchmarks test only the former. CulturalVQA \cite{nayak2024benchmarking} adds a complementary distinction between visual recognition and cultural understanding, arguing that geographically diverse datasets such as Dollar Street and GLDv2 primarily measure recognition. Tanawpamana is designed around both: it tests whether models can identify specific Philippine heritage sites, and whether they hold the associated factual, geographic, and designation knowledge.

\subsection{Talapamana Registry}
Philippine cultural heritage is governed by an established legal and institutional framework that predates any AI benchmark. Republic Act No. 10066, the National Cultural Heritage Act of 2009, as amended by Republic Act No. 11961 in 2023, provides the legal foundation for the protection and conservation of cultural heritage and strengthens the role of the NCCA as the coordinating body for heritage management \cite{ra10066, ra11961}. Through this framework, the Philippine Registry of Heritage or Talapamana was established, cataloguing cultural properties under six designation categories: National Cultural Treasures (NCT), Important Cultural Property (ICP), UNESCO World Heritage Sites, National Historical Shrines, National Historical Monuments, and National Historical Landmarks \cite{ncca2012irr}. These designations carry different levels of legal recognition and protection. Beyond national designation, international recognition provides another layer of classification. The Philippines is a State Party to the UNESCO World Heritage Convention, with six inscribed World Heritage Sites and an active Tentative List \cite{unesco_ph}.

These designation systems were developed independently of AI evaluation and reflect institutional judgments of cultural significance rather than model-derived or crowd-generated labels. By grounding Tanawpamana in this registry, site selection and categorization are anchored in institutionally recognized heritage classifications, providing a reliable foundation for dataset construction.

\section{Tanawpamana Dataset}

\textbf{Overview.} The Tanawpamana Dataset is a manually curated collection of Philippine cultural heritage site images, each paired with structured metadata describing its geographic location, heritage category, and official designation status. It is built from the NCCA's Talapamana registry \cite{talapamana}, which provides structured records of officially recognized Philippine cultural properties. As this study focuses on cultural heritage sites, the registry was filtered to include only Tangible Immovable Heritage sites in the Philippines, which we refer to hereafter as \textit{TPTI} for convenience. Tanawpamana extends TPTI into a vision-language resource by collecting representative images and standardizing and enriching the original metadata.

Designed as both a standalone resource and the foundation of the proposed Tanawpamana Benchmark (see Section~\ref{tanawpamana-bench}), the dataset's image-metadata structure can support a range of tasks beyond those evaluated in this paper, including image classification and other culturally grounded vision-language applications. It provides an authoritative representation of Philippine cultural heritage for computer vision and vision-language research.

\subsection{Data Collection}

We first filtered the Talapamana registry \cite{talapamana} to include only Tangible Immovable Heritage located in the Philippines (TPTI). For each site included in Tanawpamana, we manually gathered one representative image, with Wikimedia Commons serving as the primary image source. Only a small number of images were obtained from official government sources, namely the National Historical Commission of the Philippines (NHCP) and the National Museum of the Philippines, when no suitable Wikimedia Commons image was available. The image collection process proceeded in two stages, described below.

From TPTI, we prioritized sites with formal heritage designations, namely UNESCO World Heritage, UNESCO World Heritage Tentative List, National Cultural Treasure, and Important Cultural Property, and exhaustively collected images of these sites across all geographic areas (Metro Manila, Luzon, Visayas, and Mindanao). We refer to sites with any of these four designations as \textit{high-designation sites}, and all other sites as \textit{low-designation sites}.

After establishing this high-designation subset, we defined a set of heritage categories based primarily on each site's function, and manually assigned each high-designation site to one of these categories. We then grouped the sites by heritage category and geographic area, and identified combinations that fell short of a target of 15 sites per category–area pair, a threshold set to ensure sufficient representation for each combination.

To fill these gaps, we added low-designation sites from TPTI, collecting an image for each candidate site before including it in Tanawpamana. To do this efficiently, we used a frontier model (Claude Sonnet 5) to preliminarily assign each candidate site to a heritage category, which helped identify which sites would best fill the remaining gaps. These model-generated labels were only a rough first pass to speed up image collection. All labels were later verified and corrected manually according to the assignment rule described in Section~\ref{sec:heritage}. We also researched and standardized each site's official and alternative names as part of the dataset metadata.

All collected images went through a preprocessing procedure before being added to the dataset. First, we inspected and cropped each image to remove any post-processed text, such as labels, captions, or overlaid annotations, so that models would rely on the visual appearance of the site itself rather than on text. We also stripped embedded image metadata to remove any other non-visual information. Finally, we resized the cropped images so that neither dimension (height or width) exceeded 1024 pixels, standardizing image size and reducing computational load while preserving enough visual detail for recognition tasks.

\subsection{Feature Enrichment}
\subsubsection{Heritage Category}
\label{sec:heritage}
\begin{figure}[t]
  \centering
  \includegraphics[width=\linewidth]{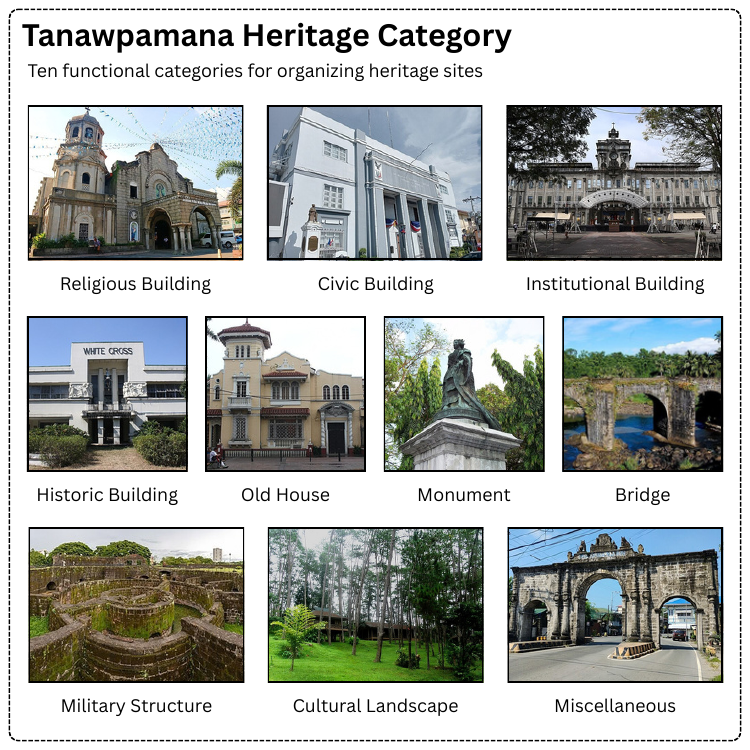}
  \caption{Heritage categories used in Tanawpamana, each shown with a sample photo.}
  \label{fig:tanawpamana_category}
\end{figure}

Since the Talapamana registry does not provide a standardized functional classification suitable for computer vision tasks, we manually defined and assigned a heritage category for each site. The final taxonomy consists of ten categories, shown with sample photos in Figure~\ref{fig:tanawpamana_category}: \textit{Religious Building}, \textit{Civic Building}, \textit{Institutional Building}, \textit{Historic Building}, \textit{Old House}, \textit{Monument}, \textit{Bridge}, \textit{Military Structure}, \textit{Cultural Landscape}, and \textit{Miscellaneous}.

These categories are not part of the official NCCA classification and were introduced solely to support benchmark construction. In particular, they facilitate the selection of visually similar distractors for the hard recognition tasks, as sites within the same heritage category often share similar visual characteristics (e.g., churches are generally more visually similar to other churches than to bridges). Because buildings constitute a large proportion of the heritage sites in the dataset, they were further subdivided into Civic Building (associated with local or national government) and Institutional Building (associated with educational institutions), with Historic Building serving as a catch-all category for buildings that fit neither classification.

As a general rule, category assignment was based primarily on each site's intended function. However, when a site's function was ambiguous, assignment instead prioritized visual appearance over institutional function, since visual similarity is what determines distractor difficulty. For example, a church affiliated with an educational institution would be classified as a Religious Building rather than an Institutional Building under this rule.
\subsubsection{Talapamana-Derived Features}

To improve Tanawpamana's suitability for vision-language research, we enriched metadata fields from Talapamana's records. These include creating individual binary flags for a site's designation statuses, such as UNESCO World Heritage, UNESCO World Heritage Tentative List, NCT, ICP, recognition as a National Artist Work, National Historical Landmark, and Heritage House, alongside expanding alternative names gathered during manual curation and standardizing site names for cases where the original name was ambiguous (e.g., ``Rizal Monument,'' which refers to any of several distinct monuments across the country).

Tanawpamana also includes a \texttt{QUESTION\_STEM} field, which reformulates each site's historical description from Talapamana into a question suitable for benchmarking. The construction of this field is described in Section~\ref{bench-cons}.

\subsubsection{Image Licensing}
\label{sec:license}

Tanawpamana includes explicit licensing information for every image, recording its source, author, license, and the modifications applied. Each image retains the license of its source: of the 499 images, 375 are licensed under CC~BY-SA, 97 are in the public domain or CC0, 19 are licensed under CC~BY, one is under an attribution-only license, and seven are Philippine government works included for non-commercial research use.

\subsection{Data Characterization}

Table~\ref{tab:category_area_counts} summarizes the distribution of the Tanawpamana Dataset across heritage categories and geographic areas. Although the collection strategy aimed to improve coverage across categories and regions, the final distribution remains non-uniform. This reflects the varying degrees to which heritage sites are documented online.

\begin{table}[h]
  \centering
  \caption{Number of sites per category and geographic area.}
  \label{tab:category_area_counts}
  \resizebox{\linewidth}{!}{%
  \begin{tabular}{lcccc}
    \toprule
    \textbf{Category} & \textbf{Metro Manila} & \textbf{Luzon} & \textbf{Visayas} & \textbf{Mindanao} \\
    \midrule
    \arrayrulecolor{highlightblue}
    \specialrule{1.2pt}{0pt}{0pt}
    Religious Building     & 21 & 42 & 23 & \lo{11} \\
    \specialrule{1.2pt}{0pt}{0pt}
    \arrayrulecolor{black}
    Civic Building         & 16 & 15 & 16 & \lo{7}  \\
    Institutional Building & 16 & 16 & 15 & \lo{4}  \\
    Historic Building      & 15 & 16 & 17 & \lo{2}  \\
    Old House              & 17 & 15 & 15 & \lo{3}  \\
    Monument               & 15 & 15 & 16 & \lo{7}  \\
    Bridge                 & \lo{7}  & 15 & \lo{4}  & \lo{2}  \\
    Military Structure     & \lo{8}  & 15 & \lo{10} & \lo{4}  \\
    Cultural Landscape     & 15 & 18 & 16 & \lo{12} \\
    Miscellaneous          & \lo{4}  & \lo{6}  & \lo{2}  & \lo{6}  \\
    \bottomrule
  \end{tabular}%
  }
  \vspace{2pt}
  \small\textcolor{softblue}{\rule{6pt}{6pt}} Below target ($<$ 15)
\end{table}

\textbf{Religious buildings are substantially overrepresented in the dataset.} As shown in Table~\ref{tab:category_area_counts}, Religious Building is the largest heritage category, comprising 97 of 499 sites (19.4\%), about 1.6 times the size of the next-largest category. This concentration is primarily driven by the exhaustive inclusion of high-designation sites, where religious heritage constitutes a substantial proportion of formally recognized properties: 89.7\% of Religious Building sites belong to the high-designation subset, compared with 41.3\% of all sites in the dataset.

\begin{figure}[h]
  \centering
  \includegraphics[width=\linewidth]{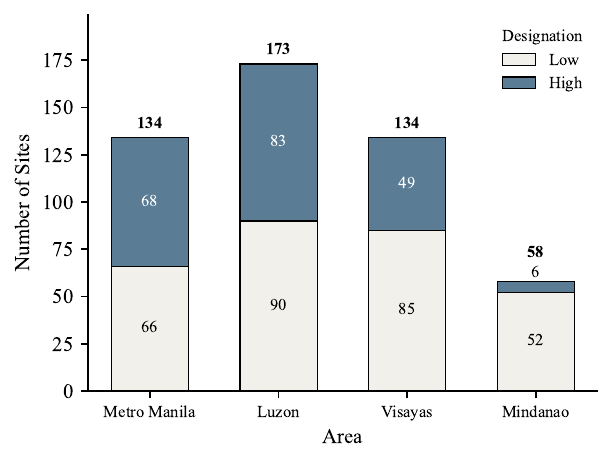}
  \caption{Distribution of heritage sites by geographic area and designation status.}
  \label{fig:area_designation}
\end{figure}

\textbf{Mindanao is underrepresented across both publicly available images and heritage designation.} Cells highlighted in soft blue in Table~\ref{tab:category_area_counts} indicate category--area combinations for which fewer than 15 images could be collected despite exhaustive search. These shortfalls occur primarily in Mindanao, suggesting limited online documentation for heritage sites in the region. Figure~\ref{fig:area_designation} further shows that Mindanao also has the lowest proportion of high-designation sites, with only 6 of 58 sites (10.3\%) belonging to the high-designation subset, compared with 50.7\%, 48.0\%, and 36.6\% for Metro Manila, Luzon, and Visayas, respectively. Together, these patterns indicate that Mindanao is disadvantaged along two dimensions: visual documentation availability and formal heritage recognition.

\section{Tanawpamana Benchmark}
\label{tanawpamana-bench}

\textbf{Overview.} The Tanawpamana Benchmark is designed to evaluate two complementary capabilities of vision-language models: (1) their ability to recognize Philippine heritage sites from images and (2) their ability to retrieve relevant cultural knowledge. Accordingly, the benchmark consists of two task suites: the Recognition Suite, which evaluates visual identification, and the Knowledge Suite, which evaluates factual, geographic, and designation-related knowledge that is largely independent of visual recognition. While both suites test only entity-level recognition and knowledge, rather than the interpretive or experiential dimensions of culture, we consider them necessary, though not sufficient, conditions for cultural understanding. A model cannot be said to understand a heritage site's cultural significance if it cannot identify the site or recall basic facts about it.

Each benchmark item is formulated as a four-option multiple-choice question. This format enables standardized evaluation across models while preserving metadata required for post-hoc analysis of performance across heritage categories, geographic areas, and designation statuses.
\subsection{Task Suite}

The Tanawpamana Benchmark consists of 4,046 multiple-choice tasks, including 1,960 Recognition Suite tasks and 2,086 Knowledge Suite tasks. A breakdown of the benchmark composition is provided in Table~\ref{tab:task_distribution}.

\begin{table}[h]
\centering
\caption{Distribution of tasks in the Tanawpamana Benchmark.}
\label{tab:task_distribution}
\begin{tabular}{l r}
\hline
\textbf{Suite} & \textbf{Task Count} \\
\hline
\multicolumn{2}{l}{\textit{Recognition Suite}} \\
\quad Image-to-Name (Easy) & 499 \\
\quad Image-to-Name (Hard) & 481 \\
\quad Name-to-Image (Easy) & 499 \\
\quad Name-to-Image (Hard) & 481 \\
\quad \textbf{Recognition Total} & \textbf{1,960} \\
\hline
\multicolumn{2}{l}{\textit{Knowledge Suite}} \\
\quad General Information & 365 \\
\quad Short Description & 481 \\
\quad Location (Text) & 499 \\
\quad Location (Image) & 499 \\
\quad Designation & 242 \\
\quad \textbf{Knowledge Total} & \textbf{2,086} \\
\hline
\textbf{Total} & \textbf{4,046} \\
\hline
\end{tabular}
\end{table}

\subsubsection{Recognition Suite}

The Recognition Suite evaluates whether models can identify Philippine heritage sites through image--name associations. It consists of two task formats: \textbf{Image-to-Name}, where a site image is provided and the model must select the corresponding heritage site name, and \textbf{Name-to-Image}, where a site name is provided and the model must identify the corresponding image.

Each format contains two difficulty levels, easy and hard. The easy setting uses cross-category distractors, where candidate sites belong to different heritage categories and can often be distinguished through broad visual characteristics. The hard setting uses same-category distractors, requiring models to distinguish between visually similar heritage sites rather than relying only on general architectural or functional cues. This design aims to evaluate fine-grained heritage recognition beyond coarse category-level recognition.

\subsubsection{Knowledge Suite}

\textbf{General Information.} The General Information task evaluates whether models can associate descriptive historical information with specific heritage sites. Given a question derived from a site's historical description, models must select the corresponding heritage site from four candidate options. This task measures the ability to retrieve cultural knowledge from contextual information such as historical events and architecture style.

\textbf{Short Description.} The Short Description task evaluates whether models can identify heritage sites from limited contextual information. Given only a heritage category and geographic area, models must select the corresponding site from four candidate options. Unlike the General Information task, which provides richer historical descriptions, this task uses coarser categorical and geographic cues, measuring whether models have broader associations among heritage sites, their function, and regional context.

\textbf{Location (Text).} The Location (Text) task evaluates geographic knowledge of heritage sites. Given the name of a heritage site, models must identify its corresponding geographic area. This task measures the ability to associate named heritage entities with their documented locations.

\textbf{Location (Image).} The Location (Image) task evaluates geographic inference from visual information. Given an image of a heritage site, models must also identify its corresponding geographic area. When compared with Location Text, this task allows analysis of the difference between explicit geographic knowledge and location inference from visual appearance.

\textbf{Designation.} The Designation task evaluates knowledge of official heritage recognition. Given a designation type, such as National Cultural Treasure or UNESCO World Heritage Site, models must identify the corresponding heritage site that holds it. This task measures whether models have knowledge of formal heritage statuses rather than relying solely on visual familiarity or general popularity.

\subsection{Benchmark Construction}
\label{bench-cons}

Except for General Information, all Tanawpamana Benchmark tasks are deterministically constructed from the TPTI Dataset. This design ensures that every item has a gold answer grounded in authoritative heritage records rather than model-generated or weakly supervised labels. General Information uses an LLM-assisted pipeline for question generation. However, it is constrained by rule-based filtering and validation, with all LLM calls performed at temperature 0 to improve reproducibility.

\textbf{Recognition Tasks.} Each heritage site was included as a target instance in the recognition tasks, ensuring that all entries appeared as correct answers. Distractors were randomly sampled from eligible candidates according to task difficulty. Easy tasks used cross-category distractors, while hard tasks used same-category distractors. The Miscellaneous category was excluded from hard recognition tasks because it did not represent a consistent visual category.

\textbf{Knowledge Tasks.} Knowledge tasks were constructed using different procedures depending on the source information used to generate each question.

For the General Information task, heritage descriptions from the TPTI Dataset were converted into question--answer pairs through an LLM-assisted pipeline. Entries without usable descriptions were first excluded. The remaining descriptions were then processed by an LLM to redact site names and alternative names while preserving informative generic terms. A second LLM acted as a judge to determine whether each redacted description contained sufficient distinguishing information for question construction. Descriptions that passed this filtering stage were converted into a question and an accompanying exclusion principle using another LLM. The exclusion principle explicitly identified the information required to distinguish the correct answer from the distractors. All LLM calls in this pipeline used GPT-4o-Mini with the temperature set to 0. The resulting 365 items underwent full manual review and were found to be specific, grounded in facts drawn from TPTI's descriptions, and to admit a single correct answer.

The Short Description task was constructed directly from each site's heritage category and geographic area. The Miscellaneous category was excluded because it did not provide a meaningful categorical constraint.

The Location Text and Image tasks were constructed from each site's geographic area metadata. Each heritage site was used as a query instance and paired with four fixed options: Metro Manila, Luzon, Visayas, and Mindanao. Location (Text) used the site name as input, while Location (Image) used the corresponding image. The identical option structure enabled direct comparison between the two task variants.

The Designation task was constructed from the designation metadata of each heritage site. For this task, we considered six designation types: UNESCO World Heritage Site, UNESCO World Heritage Tentative List, NCT, ICP, National Historical Landmark, and National Artist Work. For each site with at least one of these designations, one designation was randomly selected as the basis for question construction. Distractors were sampled from sites without the selected designation, including at least one site from the same heritage category when available.

\section{Evaluation}

\subsection{Models}

We evaluated Tanawpamana using seven vision-language models across three groups. The first group consisted of current-generation lightweight proprietary models from major model families: GPT-5.4-Mini, Gemini-3.5-Flash, and Claude-Haiku-4.5. The second group consisted of current-generation open-weight models: Qwen3.6-35B-A3B and Llama-4-Maverick. Finally, two previous-generation models, GPT-4o-Mini and Qwen3-VL-8B-Instruct, were included as legacy baselines.

All models were evaluated in a zero-shot setting using the multiple-choice format provided by the benchmark. Inference was performed with temperature set to 0. Models with optional reasoning modes were evaluated with these disabled. Instead, all models were instructed to provide a single sentence of reasoning before producing their final answer in a fixed format. Reasoning traces were not used for scoring. Only the extracted final answer choice was evaluated.

\subsection{Scoring}
Performance was measured using multiple-choice accuracy. For each question, the evaluation pipeline stored the complete model response, extracted the final predicted answer label, and compared it against the ground-truth label. The generated reasoning trace was retained separately for error analysis and was not used in determining correctness.

We also introduce two aggregate recognition scores, \textit{Recognition (Hard)} and \textit{Recognition (Easy)}, to summarize the models' performance on recognition tasks at each difficulty level. Specifically,
\[
\text{Recognition (Hard) Score} =
\frac{A_{I\rightarrow N}^{\text{Hard}} + A_{N\rightarrow I}^{\text{Hard}}}{2},
\]
where \(A_{I\rightarrow N}^{\text{Hard}}\) and \(A_{N\rightarrow I}^{\text{Hard}}\) denote the accuracies of the Image-to-Name (Hard) and Name-to-Image (Hard) tasks, respectively. Similarly, Recognition (Easy) Score is defined as
\[
\text{Recognition (Easy) Score} =
\frac{A_{I\rightarrow N}^{\text{Easy}} + A_{N\rightarrow I}^{\text{Easy}}}{2},
\]
where \(A_{I\rightarrow N}^{\text{Easy}}\) and \(A_{N\rightarrow I}^{\text{Easy}}\) are the accuracies of the Image-to-Name (Easy) and Name-to-Image (Easy) tasks, respectively.

\subsection{Error Parsing}
Responses that did not contain a valid answer in the required format, with a maximum output length of 1,024 tokens, were considered incorrect. Manual inspection of these parsing failures revealed that most resulted from models indicating that the correct answer was not among the provided options, or from producing overly verbose responses that prevented reliable extraction of the final answer.

\section{Results}
\label{results}
\begin{figure*}[h]
  \centering
  \includegraphics[width=0.84\textwidth]{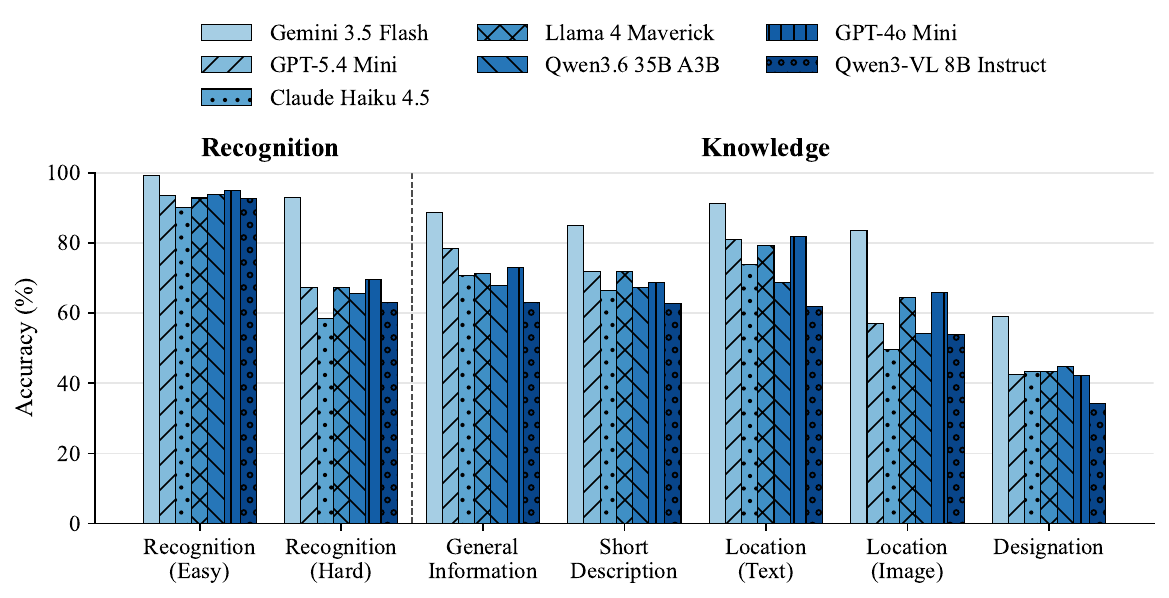}
  \caption{Model accuracy across seven task clusters spanning the
Recognition (Easy/Hard) and five Knowledge tasks.}
  \label{fig:main_results}
\end{figure*}

Figure~\ref{fig:main_results} summarizes model performance across the Recognition and Knowledge suites. Overall, models perform well above chance (25\%, or 34.7\% for the Location tasks; see Section~\ref{bench-cons}) on nearly all tasks, demonstrating varying levels of capability in identifying and reasoning about Philippine heritage sites. Within the Recognition track, models perform substantially better on Easy tasks than on Hard tasks. The lowest performance overall is observed on the Designation task, which requires updated knowledge of formal heritage status.

\subsection{Recognition}

\textbf{Gemini-3.5-Flash consistently achieves the strongest recognition performance.} As shown in Figure~\ref{fig:main_results}, Gemini-3.5-Flash attains the highest accuracy on both Recognition tasks, substantially outperforming the remaining models. The performance gap is modest on Recognition (Easy), where all models score between 90.0\% and 99.2\%, but widens considerably on Recognition (Hard): Gemini-3.5-Flash achieves 92.9\% accuracy, compared to a next-best score of 69.4\%, indicating a substantially greater ability to distinguish visually similar heritage sites.

\textbf{Recognition (Hard) provides substantially greater discriminative power than Recognition (Easy).} Performance on Recognition (Easy) is largely saturated, spanning only 9.2 percentage points across all evaluated models (90.0--99.2\%). In contrast, Recognition (Hard) spans 34.4 percentage points (58.5--92.9\%), producing much greater separation between models. Gemini-3.5-Flash exhibits only a 6.3-point drop from Easy to Hard, whereas every other model declines by approximately 25.6--31.5 points. Despite this divergence in difficulty, per-model accuracy on Easy and Hard is strongly correlated ($r=0.95$), indicating that Recognition (Hard) preserves much of the same overall model ranking while substantially amplifying the differences between them. These results suggest that the same-category distractor construction successfully transforms the task from coarse landmark recognition into fine-grained visual discrimination, making Recognition (Hard) the primary source of discriminative signal within the benchmark.

\textbf{Recognition accuracy is largely invariant across geographic areas.} Despite substantial differences in heritage-site availability across the four geographic areas represented in the dataset, Recognition (Hard) performance remains remarkably consistent. Average accuracy ranges from 66.7\% in Luzon to 72.3\% in the Visayas, with Metro Manila (68.8\%) and Mindanao (69.1\%) performing comparably to the remaining areas with no detectable disadvantage. The same pattern holds, even more narrowly, on Recognition (Easy) (91.6--95.2\%). These findings suggest that recognition performance does not track an area's underlying documentation scarcity, though we return to this finding with appropriate caution in Section~\ref{sec:disc-gap} given Mindanao's comparatively fewer sites.

\textbf{High-designation heritage sites are more difficult to recognize.} Recognition (Hard) score is unexpectedly lower for high-designation sites than for low-designation sites, at 61.9\% and 74.1\%, respectively, a gap of 12.3 percentage points. However, it is important to note that high- and low-designation sites are not evenly distributed across heritage categories, introducing a potential confounding factor. For example, Religious Building sites comprise 44.2\% of the high-designation group but only 3.5\% of the low-designation group, and this category is comparatively difficult to recognize regardless of designation status (55.2\% average accuracy across all models, the third-lowest among all categories). Consequently, part of the observed 12.3-point gap may reflect differences in category composition rather than designation status itself. To quantify this effect, we recalculated the overall accuracy of each designation group after reweighting category-specific accuracies so that both groups shared the same heritage-category distribution. Instead of using each group's actual category proportions, the same set of category weights was applied to both groups. Under this standardization, high-designation accuracy increases to 66.7\% and low-designation accuracy decreases to 72.4\%, reducing the gap from 12.3 to 5.7 percentage points. Thus, 6.6 of the original 12.3-point gap (53.7\%) is attributable to differences in category composition, while the remaining 5.7-point gap persists after controlling for category.

\subsection{Knowledge}

\textbf{Gemini-3.5-Flash consistently achieves the strongest knowledge performance.} It attains the highest accuracy on all five Knowledge tasks, with a substantial lead over every other model. Among the remaining models, however, proprietary models do not consistently outperform open-weight ones. Averaged across the five tasks, the open-weight Llama-4-Maverick (66.0\%) performs on par with GPT-5.4-Mini (66.1\%) and clearly exceeds Claude-Haiku-4.5 (60.8\%), while the open-weight Qwen3.6-35B-A3B (60.6\%) is on par with Claude-Haiku-4.5. Knowledge performance therefore depends more on the individual model than on whether it is proprietary or open-weight.

\textbf{Heritage designation is the most challenging knowledge task.} Designation proves difficult for every evaluated model, with accuracy ranging from only 34.3\% to 59.1\% and only Gemini-3.5-Flash scoring above 50\%. This contrasts sharply with tasks such as General Information and Location (Text), where several models achieve substantially higher accuracy. The consistency of this pattern across all model families suggests that heritage designation is a distinct and comparatively difficult form of cultural knowledge for current vision-language models. Part of this difficulty may stem from designation status being periodically updated, which requires models to have current knowledge of a site's official recognition.

\textbf{Models exhibit substantially stronger geographic reasoning from text than from images.} The Location (Text) and Location (Image) tasks test the same knowledge, the geographic area of a heritage site, and differ only in whether the site is given as its name or as an image. Every evaluated model loses accuracy when given an image, with drops ranging from 7.6 percentage points for Gemini-3.5-Flash (91.2\% to 83.6\%) and 7.8 for Qwen3-VL-8B-Instruct to 24.0 for both GPT-5.4-Mini and Claude-Haiku-4.5. This consistent gap indicates that these models often possess the relevant geographic knowledge but fail to reliably ground it when given only an image.    

\section{Discussion}

\subsection{Regional Representation Gaps}
\label{sec:disc-gap}

Mindanao's representational scarcity in Tanawpamana, both in publicly available images and in high-designation heritage sites, reflects a broader pattern documented in Philippine scholarship. The region remains comparatively low-resource for both natural language and vision-language research, and has been described in cultural scholarship as ``relegated in academic research and scholarship to peripheral status at best, silence more commonly... total erasure at worst'' relative to Luzon-centered narratives of national culture~\cite{paredes2022making}.

Such geographic imbalances are generally associated with reduced recognition performance in machine learning datasets~\cite{de2019does}. We find no evidence of this in Tanawpamana: on Recognition (Hard), Mindanao sites and sites from the remaining areas both average 69.1\%. This should not be read as regional parity, however, for two reasons.

First, the comparison has limited statistical power. Mindanao contributes only 52 sites to the Recognition (Hard) tasks, compared with 429 from the remaining areas. Because all seven models are evaluated on the same sites, their responses to a given site are not independent, so we use the site as the unit of analysis, scoring each site by its accuracy over its 14 Recognition (Hard) responses (two tasks across seven models). Resampling sites with replacement 10,000 times gives a 95\% bootstrap confidence interval of $-7.7$ to $+7.5$ percentage points for the Mindanao difference. A two-sample comparison of these group sizes would detect a true difference with 80\% power ($\alpha = 0.05$) only if it were about 11 percentage points or larger, given a site-level standard deviation of 27.7 percentage points. A Mindanao disadvantage larger than about 8 percentage points is therefore unlikely, but a smaller one cannot be excluded.

Second, Mindanao sites may be unrepresentative. Because Tanawpamana relies exclusively on publicly available imagery, the collected Mindanao sites may be biased toward those with disproportionately strong online documentation, and thus a greater likelihood of prior model exposure. The observed recognition performance may therefore reflect the visibility of the sites that passed our collection criteria rather than genuine regional parity in model knowledge.

Taken together, our results show an absence of evidence for a large Mindanao disadvantage, not evidence that models recognize Mindanao heritage as well as heritage elsewhere. Settling this question would require more Mindanao sites, including less-documented ones.

\subsection{Fine-Grained Discrimination as a Bottleneck}

Every model performs worse on Recognition (Hard) than on Recognition (Easy), showing the difficulty introduced by same-category distractors that look similar to the target. This supports the distinction drawn by OVEN~\cite{hu2023open} between category-level and entity-level recognition: models that fail on Tanawpamana's hard tasks typically confuse the target with another site of the same category rather than failing at recognition entirely. FG-BMK~\cite{yu2025benchmarking} similarly finds that current large vision-language models remain limited fine-grained recognizers across domains. Our results show the same pattern for Philippine heritage, which is that models often identify a site's general category correctly but fail to identify the specific site.

Gemini-3.5-Flash's much smaller Easy-to-Hard drop suggests that its advantage lies in more precise visual discrimination rather than broader landmark familiarity. Overall, the key challenge for current VLMs in heritage recognition is not identifying general architectural types, but distinguishing between visually similar sites.

\subsection{The Gap Between Visual and Textual Geographic Knowledge}
 
The gap between text-based and image-based geographic knowledge aligns with broader findings in multimodal evaluation, where vision-language models often fail to match text-only models on knowledge that is nominally associated with the visual world. A recent cross-modal benchmark, for example, found that VLMs perform significantly worse on visual grounding dimensions compared with other modalities representing the same underlying knowledge~\cite{yang2025does}.

Our results extend this observation to a real-world cultural and geographic setting. The same geographic fact (i.e., a heritage site's area) is retrieved substantially more reliably when provided with the site's name than when provided only with an image. This suggests that much of current VLMs' geographic knowledge remains encoded through textual associations rather than grounded visual representations. Improving this modality alignment remains an open challenge distinct from increasing a model's factual knowledge capacity.

\subsection{Error Analysis}
\label{sec:error-analysis}
We manually analyzed errors from Gemini-3.5-Flash, the best-performing model among those evaluated, using sampled model reasoning across different benchmark tasks. We identify the most common failure patterns observed across recognition and knowledge categories.

In Recognition (Easy), predictions were primarily based on broad architectural reasoning. The model relied on visible characteristics such as building style and structure, which were usually sufficient for distinguishing distractors. In Recognition (Hard), however, the same visual reasoning strategy became less effective due to same-category distractors. As a result, errors in this task were dominated by \textit{visually similar site confusion}. The model continued to rely on architectural features, but these often lacked the discriminative power needed to distinguish sites within the same heritage category, leading it to predict visually similar yet incorrect candidates.

Across General Information, Short Description, and Designation, the dominant failure mode was \textit{misattribution}, that is, the model associated the provided historical information, description, or designation status with an incorrect heritage site among the candidate answers. A similar pattern appeared in Location (Text), where errors typically involved assigning a heritage site to the wrong geographic area.

Unlike these text-based tasks, Location (Image) errors were primarily caused by \textit{visual misidentification}. The model first confused the depicted heritage site with another candidate and subsequently selected the geographic information associated with that incorrectly identified site.

In summary, recognition failures primarily reflect difficulties in fine-grained visual discrimination, whereas knowledge failures more often arise from incorrect associations between heritage entities and their corresponding attributes.

\section{Limitations}
\label{sec:limitations}

\subsection{Tanawpamana Dataset}

\textbf{The heritage categories introduced in Tanawpamana are benchmark-oriented rather than official classifications.} The NCCA Talapamana registry does not provide a functional taxonomy for computer vision applications, so we introduced categories based on site function and visual characteristics. Some sites' category assignment is inherently ambiguous. These categories should therefore be interpreted as benchmark-oriented groupings rather than as formal cultural classifications.


\textbf{The dataset reflects a snapshot of the Talapamana registry.} The registry was accessed in January 2026, and since Talapamana is periodically updated, including updates recorded in May 2026, Tanawpamana represents a time-stamped snapshot rather than a permanent inventory of Philippine heritage records. This is most consequential for the dataset's designation fields, which are the metadata most likely to change over time and may no longer reflect a site's current status.

\subsection{Tanawpamana Benchmark}

\textbf{Each heritage site is represented by only one image.} Recognition performance may therefore partly reflect image-specific factors, such as viewpoint, lighting, resolution, or framing, rather than a model's actual familiarity with the site. Future work can expand the dataset with additional images per site to separate these factors from site-level recognition, and further verify the collected images.


\textbf{The evaluation does not include regionally adapted models.} All evaluated models are general-purpose. Models trained with greater Southeast Asian coverage, such as SEA-LION~\cite{ng2025sea} and Aya Vision~\cite{dash2025aya}, could be evaluated in future work to test whether regional training improves recognition and knowledge of Philippine heritage.


\textbf{Heritage category assignment was performed by a single researcher.} While assignment followed a defined, function-based rule rather than ad hoc judgment, inter-annotator agreement was not assessed. Since Recognition (Hard) distractors are sampled based on these categories, task difficulty may partly reflect category design choices in addition to models' recognition ability. Future work could quantify agreement between independent annotators applying this categorization.

\textbf{The Designation task is particularly time-sensitive and may become outdated for future model comparisons.} Because official designation status can change as the Talapamana registry is updated, gold labels for this task may no longer be accurate over time. We recommend that future use of the benchmark either refresh Designation labels against the current registry or exclude this task from comparisons.

\textbf{The benchmark does not include a human performance baseline.} It therefore remains unclear how model performance compares to human recognition and knowledge.

\section{Conclusion}

In this paper, we introduce the Tanawpamana Dataset, a collection of images of Philippine heritage sites, with additional information derived from the NCCA's Talapamana registry, along with the Tanawpamana Benchmark, an evaluation suite designed to assess both visual recognition of Philippine heritage sites and cultural knowledge about them. The dataset itself reflects the current state of Philippine heritage documentation, with Mindanao remaining underrepresented compared to other geographic areas.

From our benchmarking, we found that recognition accuracy across all evaluated models remains high when distractors belong to different heritage categories. However, performance declines sharply when models must distinguish visually similar sites within the same category, indicating that current models struggle with fine-grained, entity-level discrimination rather than coarse visual recognition. Furthermore, a separate pattern emerges for geographic knowledge, with models consistently inferring a site's geographic area more accurately from its name than from its image. This finding suggests that geographic knowledge is more reliably recalled from text than from visual input alone.

The findings of this study highlight important limitations in current vision-language models when recognizing and reasoning about culturally specific heritage sites. They also demonstrate the value of evaluating both visual recognition and cultural knowledge, providing a more comprehensive understanding of model capabilities on underrepresented cultural domains such as Philippine cultural heritage sites.

\clearpage
\section{Acknowledgments}
This study was supported by JSPS KAKENHI Grant Numbers JP26H02537 and JP25KJ1833.
This study was extended from a project conducted under the NAIST Project for Interns (NAPI), a partnership between the Nara Institute of Science and Technology and Ateneo de Manila University. We express our gratitude to the teams who organized and supported the program.
\noindent\textbf{Photo credits.} Photos in Figures~1 and~2 are from Wikimedia Commons, cropped and resized by the authors.
Fig.~1: Ian Jay Traquena (CC~BY-SA~3.0).
Fig.~2, CC~BY-SA~4.0: Diosa Marie Aguila (Religious Building), Ralff Nestor Nacor (Civic Building, Miscellaneous), Patrickroque01 (Institutional Building), Diego Delso (Military Structure);
CC~BY-SA~3.0: Claire Algarme (Historic Building), Elmer B.\ Domingo (Old House), Dondie Garcia (Monument), Carol Zornosa (Bridge), Bayronnoel (Cultural Landscape).
Source URLs are listed in \texttt{dataset/Tanawpamana\_license.csv} on GitHub.

\section{Generative AI Use Disclosure}
AI language models (Claude) were used for proofreading and editing manuscript text. Claude Code was used as a coding assistant. All quantitative results, claims, and figures were independently verified by the authors. No AI-generated content was used for research design or analysis unless otherwise stated in the paper.

\bibliographystyle{IEEEtran}
\bibliography{mybib}

@article{pranav2025rice,
  title={{Rice-VL}: Evaluating Vision-Language Models for Cultural Understanding Across {ASEAN} Countries},
  author={Pranav, Tushar and Pandey, Eshan and Bala, Austria Lyka Diane and Chadha, Aman and Atmosukarto, Indriyati and Lock, Donny Soh Cheng},
  journal={arXiv preprint arXiv:2512.01419},
  year={2025}
}

@inproceedings{lovenia2024seacrowd,
  title={{SEACrowd}: A Multilingual Multimodal Data Hub and Benchmark Suite for Southeast Asian Languages},
  author={Lovenia, Holy and Mahendra, Rahmad and Akbar, Salsabil Maulana and Miranda, Lester James Validad and Santoso, Jennifer and Aco, Elyanah and Fadhilah, Akhdan and Mansurov, Jonibek and Imperial, Joseph Marvin and Kampman, Onno P. and others},
  booktitle={Proceedings of the 2024 Conference on Empirical Methods in Natural Language Processing},
  pages={5155--5203},
  year={2024}
}

@inproceedings{cahyawijaya2025crowdsource,
  title={{Crowdsource, Crawl, or Generate? Creating {SEA-VL}, a Multicultural Vision-Language Dataset for Southeast Asia}},
  author={Cahyawijaya, Samuel and Lovenia, Holy and Moniz, Joel Ruben Antony and Wong, Tack Hwa and Farhansyah, Mohammad Rifqi and Maung, Thant Thiri and Hudi, Frederikus and Anugraha, David and Habibi, Muhammad Ravi Shulthan and Qorib, Muhammad Reza and others},
  booktitle={Proceedings of the 63rd Annual Meeting of the Association for Computational Linguistics (Volume 1: Long Papers)},
  pages={18685--18717},
  year={2025}
}

@inproceedings{satar2025seeing,
  title={{Seeing Culture}: A Benchmark for Visual Reasoning and Grounding},
  author={Satar, Burak and Ma, Zhixin and Irawan, Patrick Amadeus and Mulyawan, Wilfried Ariel and Jiang, Jing and Lim, Ee-Peng and Ngo, Chong-Wah},
  booktitle={Proceedings of the 2025 Conference on Empirical Methods in Natural Language Processing},
  pages={22238--22254},
  year={2025}
}

@inproceedings{urailertprasert2024sea,
  title={{SEA-VQA}: Southeast Asian Cultural Context Dataset for Visual Question Answering},
  author={Urailertprasert, Norawit and Limkonchotiwat, Peerat and Suwajanakorn, Supasorn and Nutanong, Sarana},
  booktitle={Proceedings of the 3rd Workshop on Advances in Language and Vision Research ({ALVR})},
  pages={173--185},
  year={2024}
}

@inproceedings{antol2015vqa,
  title={{VQA}: Visual Question Answering},
  author={Antol, Stanislaw and Agrawal, Aishwarya and Lu, Jiasen and Mitchell, Margaret and Batra, Dhruv and Zitnick, C. Lawrence and Parikh, Devi},
  booktitle={Proceedings of the IEEE International Conference on Computer Vision},
  pages={2425--2433},
  year={2015}
}

@inproceedings{yue2024mmmu,
  title={{MMMU}: A Massive Multi-discipline Multimodal Understanding and Reasoning Benchmark for Expert {AGI}},
  author={Yue, Xiang and Ni, Yuansheng and Zhang, Kai and Zheng, Tianyu and Liu, Ruoqi and Zhang, Ge and Stevens, Samuel and Jiang, Dongfu and Ren, Weiming and Sun, Yuxuan and others},
  booktitle={Proceedings of the IEEE/CVF Conference on Computer Vision and Pattern Recognition},
  pages={9556--9567},
  year={2024}
}

@inproceedings{hu2023open,
  title={{Open-Domain Visual Entity Recognition}: Towards Recognizing Millions of Wikipedia Entities},
  author={Hu, Hexiang and Luan, Yi and Chen, Yang and Khandelwal, Urvashi and Joshi, Mandar and Lee, Kenton and Toutanova, Kristina and Chang, Ming-Wei},
  booktitle={Proceedings of the IEEE/CVF International Conference on Computer Vision},
  pages={12065--12075},
  year={2023}
}

@inproceedings{nayak2024benchmarking,
  title={{Benchmarking Vision Language Models for Cultural Understanding}},
  author={Nayak, Shravan and Jain, Kanishk and Awal, Rabiul and Reddy, Siva and Van Steenkiste, Sjoerd and Hendricks, Lisa Anne and Stanczak, Karolina and Agrawal, Aishwarya},
  booktitle={Proceedings of the 2024 Conference on Empirical Methods in Natural Language Processing},
  pages={5769--5790},
  year={2024}
}

@misc{ra10066,
  author       = {{Congress of the Philippines}},
  title        = {Republic Act No. 10066: National Cultural Heritage Act of 2009},
  year         = {2009},
  howpublished = {\url{https://lawphil.net/statutes/repacts/ra2010/ra_10066_2010.html}}
}

@misc{ra11961,
  author       = {{Congress of the Philippines}},
  title        = {Republic Act No. 11961: An Act Strengthening the Conservation and Protection of Philippine Cultural Heritage through Cultural Mapping and an Enhanced Cultural Heritage Education Program, Amending for the Purpose Republic Act No. 10066},
  howpublished = {\url{https://lawphil.net/statutes/repacts/ra2023/ra_11961_2023.html}},
  year         = {2023}
}

@misc{ncca2012irr,
  author       = {{Republic of the Philippines}},
  title        = {{Implementing Rules and Regulations of Republic Act No. 10066, Otherwise Known as the National Cultural Heritage Act of 2009}},
  year         = {2012},
  howpublished = {\url{https://www.wipo.int/wipolex/en/legislation/details/13753}}
}

@misc{unesco_ph,
  author       = {{UNESCO World Heritage Centre}},
  title        = {{Philippines} -- States Parties},
  howpublished = {\url{https://whc.unesco.org/en/statesparties/ph}},
  note         = {Accessed: 2026-01-30}
}

@misc{talapamana,
  author       = {{National Commission for Culture and the Arts}},
  title        = {{TALAPAMANA}: Philippine Registry of Heritage -- Cultural Property Database},
  howpublished = {\url{https://talapamana.ncca.gov.ph/index.php/talapamana/cultural-property-database/talapamana}},
  note         = {Accessed: Jan. 2026. Registry subsequently updated as of May 2026}
}

@article{paredes2022making,
  author    = {Oona Paredes},
  title     = {Making Mindanao: Place-Making and People-Making in the Southern Philippines},
  journal   = {South East Asia Research},
  volume    = {30},
  number    = {1},
  pages      = {3--8},
  year       = {2022},
  publisher  = {Taylor \& Francis}
}

@inproceedings{de2019does,
  author    = {Terrance De Vries and Ishan Misra and Changhan Wang and Laurens van der Maaten},
  title     = {Does Object Recognition Work for Everyone?},
  booktitle = {Proceedings of the {IEEE/CVF} Conference on Computer Vision and Pattern Recognition Workshops},
  pages      = {52--59},
  year       = {2019}
}

@misc{yu2025benchmarking,
  author        = {Hong-Tao Yu and Yuxin Peng and Serge Belongie and Xiu-Shen Wei},
  title         = {Benchmarking Large Vision-Language Models on Fine-Grained Image Tasks: A Comprehensive Evaluation},
  year          = {2025},
  eprint        = {2504.14988},
  archivePrefix = {arXiv},
  primaryClass  = {cs.CV}
}

@inproceedings{yang2025does,
  author    = {Zhihui Yang and Yupei Wang and Kaijie Mo and Zhe Zhao and Renfen Hu},
  title     = {Does Visual Grounding Enhance the Understanding of Embodied Knowledge in Large Language Models?},
  booktitle = {Findings of the Association for Computational Linguistics: {EMNLP} 2025},
  pages     = {16960--16978},
  year      = {2025}
}

@inproceedings{maulana2025limvo,
  title        = {Limvo: A Less Is More Approach for Visual Reasoning in Knowledge-Based Visual Question Answering},
  author       = {Maulana, Ade Rohmat and Suryani, Arie Ardiyanti and Rachmawati, Ema},
  booktitle    = {2025 28th Conference of the Oriental COCOSDA International Committee for the Co-ordination and Standardisation of Speech Databases and Assessment Techniques (O-COCOSDA)},
  pages        = {1--6},
  year         = {2025},
  organization = {IEEE}
}

@inproceedings{ng2025sea,
  title={{SEA-LION}: {S}outheast {A}sian Languages in One Network},
  author={Ng, Raymond and Nguyen, Thanh Ngan and Huang, Yuli and Tai, Ngee Chia and Leong, Wai Yi and Leong, Wei Qi and Yong, Xianbin and Ngui, Jian Gang and Susanto, Yosephine and Cheng, Nicholas and others},
  booktitle={Proceedings of the 14th International Joint Conference on Natural Language Processing and the 4th Conference of the Asia-Pacific Chapter of the Association for Computational Linguistics},
  pages={512--526},
  year={2025}
}

@article{dash2025aya,
  title={{Aya Vision}: Advancing the Frontier of Multilingual Multimodality},
  author={Dash, Saurabh and Nan, Yiyang and Dang, John and Ahmadian, Arash and Singh, Shivalika and Smith, Madeline and Venkitesh, Bharat and Shmyhlo, Vlad and Aryabumi, Viraat and Beller-Morales, Walter and others},
  journal={arXiv preprint arXiv:2505.08751},
  year={2025}
}

\end{document}